\documentclass[conference]{IEEEtran}
\IEEEoverridecommandlockouts
\usepackage{cite}
\usepackage{amsmath,amssymb,amsfonts}
\usepackage{algorithmic}
\usepackage{graphicx}
\usepackage{balance}
\usepackage{textcomp}
\usepackage{xcolor}
\usepackage{subfigure}
\usepackage{makecell}

\def\BibTeX{{\rm B\kern-.05em{\sc i\kern-.025em b}\kern-.08em
    T\kern-.1667em\lower.7ex\hbox{E}\kern-.125emX}}

\DeclareRobustCommand*{\IEEEauthorrefmark}[1]{%
    \raisebox{0pt}[0pt][0pt]{\textsuperscript{\footnotesize\ensuremath{#1}}}}

\begin{document}

\title{Integrated and Lightweight Design of\\ 
Electro-hydraulic Ankle Prosthesis}

\author{
\IEEEauthorblockN{
Yi Wei\IEEEauthorrefmark{1,2},
Xingjian Wang\IEEEauthorrefmark{1,2,3*},
Xinyu Tian\IEEEauthorrefmark{2}, 
Shaoping Wang\IEEEauthorrefmark{1,2,3},
Rujun Jia\IEEEauthorrefmark{2}
}
\IEEEauthorblockA{\IEEEauthorrefmark{1}Tianmushan Laboratory, Xixi Octagon City, Yuhang District, Hangzhou 310023, China}
\IEEEauthorblockA{\IEEEauthorrefmark{2}School of Automation Science and Electrical Engineering, Beihang University, Beijing 100191, China}
\IEEEauthorblockA{\IEEEauthorrefmark{3}Ningbo Institute of Technology, Beihang University, Ningbo 315800, China}
\IEEEauthorblockA{*Email: $wangxj@buaa.edu.cn$}
}

\maketitle

\begin{abstract}
For lower limb amputees, an active ankle joint prosthesis can provide basic mobility functions. This study focuses on an ankle joint prosthesis system based on the principle of electric-hydraulic actuation. By analyzing the characteristics of human gait cycles and the mechanics of ankle joint movement, a lightweight and integrated ankle joint prosthesis is designed, considering the requirements for normal ankle joint kinematics and dynamics. The components of the prosthesis are optimized through simulation and iterative improvements, while ensuring tight integration within minimal space. The design and simulation verification of the integrated lightweight prosthesis components are achieved. This research addresses the contradiction between the high output capability and the constraints on volume and weight in prosthetic devices.
\end{abstract}

\section{Introduction}
For individuals with below-knee amputations, an ankle-foot prosthesis can fulfill their basic mobility functions due to the ankle joint is one of the most important weight-bearing joints in the lower limbs of the human body. 60\% of the force that pushes the body forward and upward is generated by the ankle joint. For lower leg amputees, an ankle prosthesis can effectively replace the residual foot and restore basic mobility. Ankle prostheses can currently be divided into three categories: passive ankle prostheses, semi-active ankle prostheses, and active ankle prostheses. Among them, the active ankle prosthesis can adjust the overall damping, actively generate the power required for human movement, and reduce fatigue. It is more similar to natural limbs and thus has been extensively studied \cite{b1}.At present, the mainstream active ankle prostheses are divided into motor direct drive type represented by MIT \cite{b2} and micro-hydraulic actuated type represented by the University of Bath \cite{b3}.\\
\indent However, currently available active ankle-foot prostheses still have significant limitations. Firstly, as wearable devices, active ankle-foot prostheses are naturally constrained by size and weight, which requires the prostheses to be compact and portable while fulfilling the functional substitution of a normal human ankle joint. Secondly, existing active ankle-foot prostheses suffer from significant energy wastage, resulting in low energy utilization efficiency. This compels the prostheses to rely on larger power sources, batteries, and other components to meet the required power and usage time, contradicting the essential requirement for prostheses to be lightweight and compact.\\
\indent Therefore, this paper aims to design an integrated lightweight active ankle-foot prosthesis based on electro-hydraulic actuation. The proposed design focuses on maximizing energy utilization through energy-domain allocation and minimizing energy losses. A hydraulic integration block is utilized to tightly connect all components within the smallest possible space, enabling the lightweight design of the prosthesis components. The ultimate goal is to develop an integrated lightweight active electro-hydraulic ankle-foot prosthesis.

\section{Analysis of biomechanical parameters and modeling of electro-hydraulic mechanisms}

\subsection{Analysis of ankle joint biomechanical parameters}

Human walking is characterized by cyclic movements of the body, and therefore, gait cycles can be used to describe the features of walking. As shown in Figure 1, in a normal walking pattern, a gait cycle begins with the heel strike (HS) when the foot contacts the ground and ends with the subsequent heel strike of the same foot. A gait cycle can be divided into four phases based on the angle of the foot and its contact with the ground.:Controlled Plantar Flexion (CP),Controlled Dorsiflexion (CD),Powered Plantar Flexion (PP):and Swing (SW).

\begin{figure}[htbp]
\centerline{\includegraphics[width=0.8\linewidth]{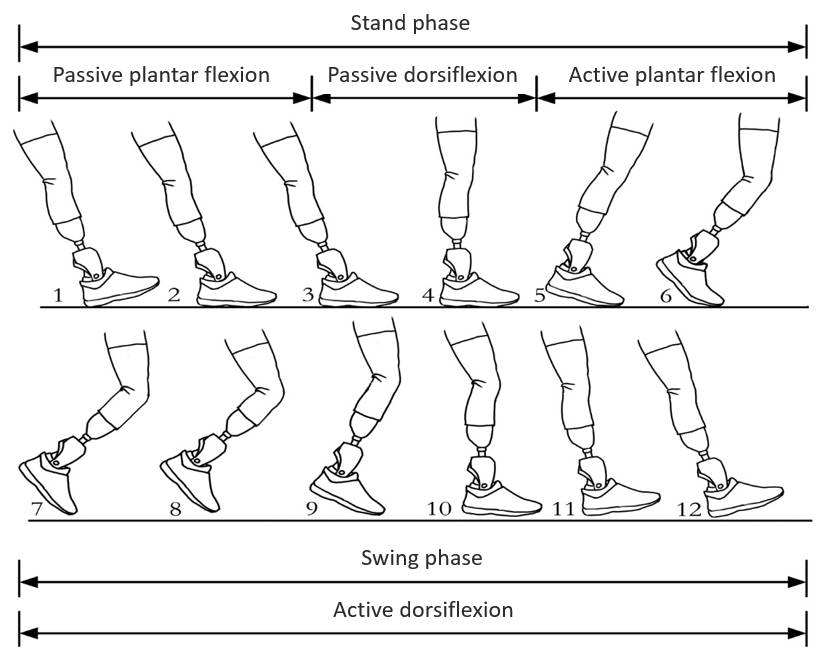}}
\caption{Illustration of Human Gait Cycle}
\end{figure}

\indent After delineating the gait cycle, the following experiment was designed to investigate the kinematics and kinetics of the ankle joint. As shown in Figure 2, the VICON motion capture system was employed to capture and record the motion data of the lower limb markers on four participants. Additionally, electromyographic (EMG) and plantar pressure signals were collected. Each participant underwent three sets of experiments, ensuring uniformity by maintaining the same speed on the treadmill throughout the entire experiment.Upon obtaining the experimental results, the gait characteristic curves of different subjects were averaged, resulting in the output characteristics of the human ankle joint during the walking process.

\begin{figure}[htbp]
\centerline{\includegraphics[width=0.4\linewidth]{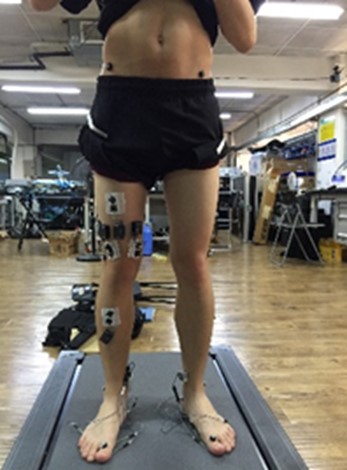}}
\caption{Ankle motion information acquisition experimental sensor}
\label{fig}
\end{figure}

\indent Figure 3 shows the ankle joint angle, torque, and power curves during one gait cycle, where the negative segment in the power curve represents the human body's work on the ankle joint and energy storage completion. The cycle starts from heel contact with the ground. Based on experimental information, it can be observed that the ankle joint prosthesis angle variation corresponds to the four-step gait and meets the requirements for peak power and peak torque. Specifically, it is noted that for most of the gait cycle, the power is below 10\% of the peak power. If direct actuation is used for the functionality through the driving mechanism, it would demand a higher output capacity for the mechanism, resulting in excessive weight and volume. Therefore, a reasonable control and configuration of the overall energy of the system are required.

\begin{figure}[htbp]
    \begin{minipage}[t]{0.32\linewidth}
        \centering
        \includegraphics[width=\textwidth]{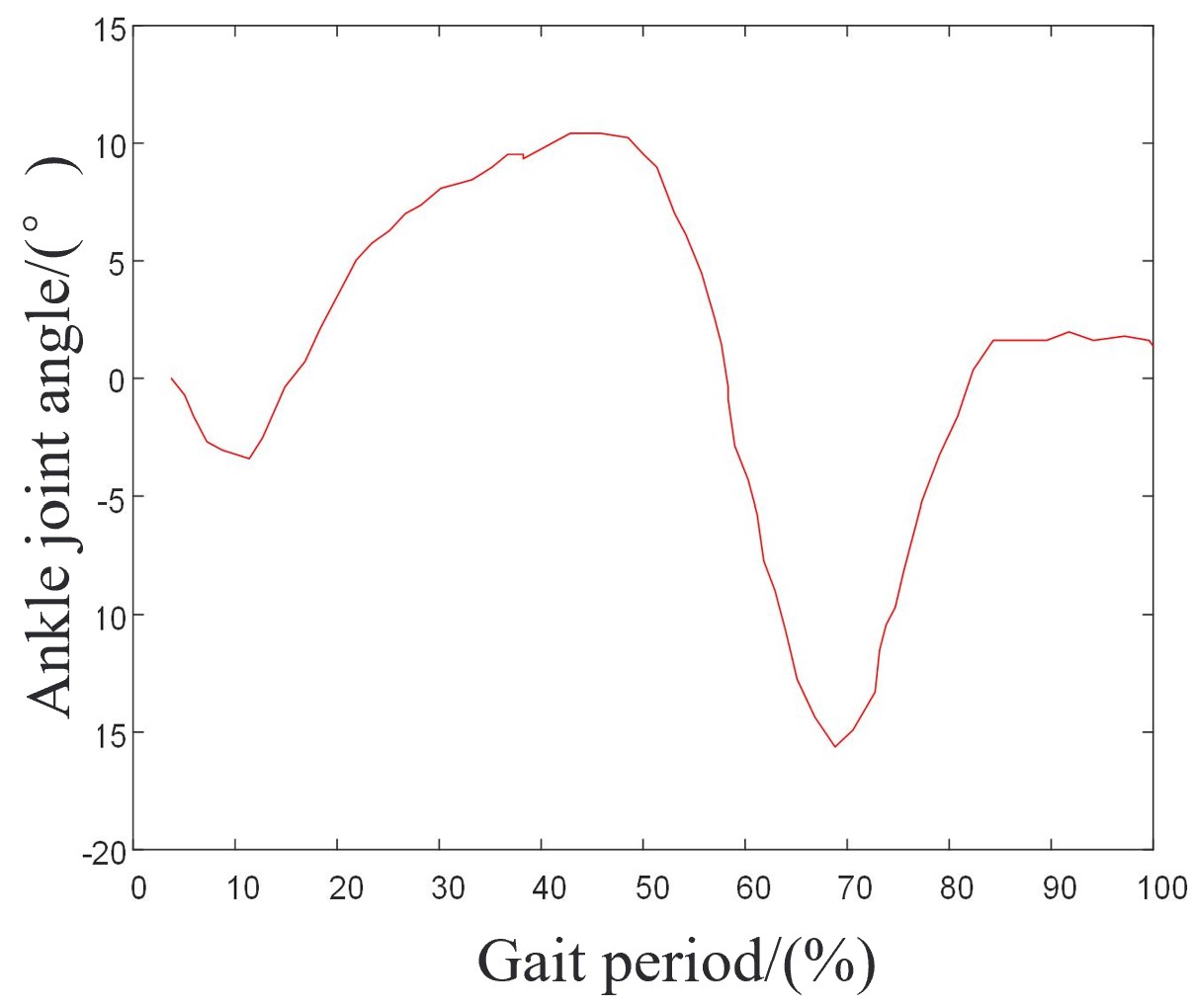}
        \centerline{\small(a)Ankle Angle}
    \end{minipage}%
    \begin{minipage}[t]{0.32\linewidth}
        \centering
        \includegraphics[width=\textwidth]{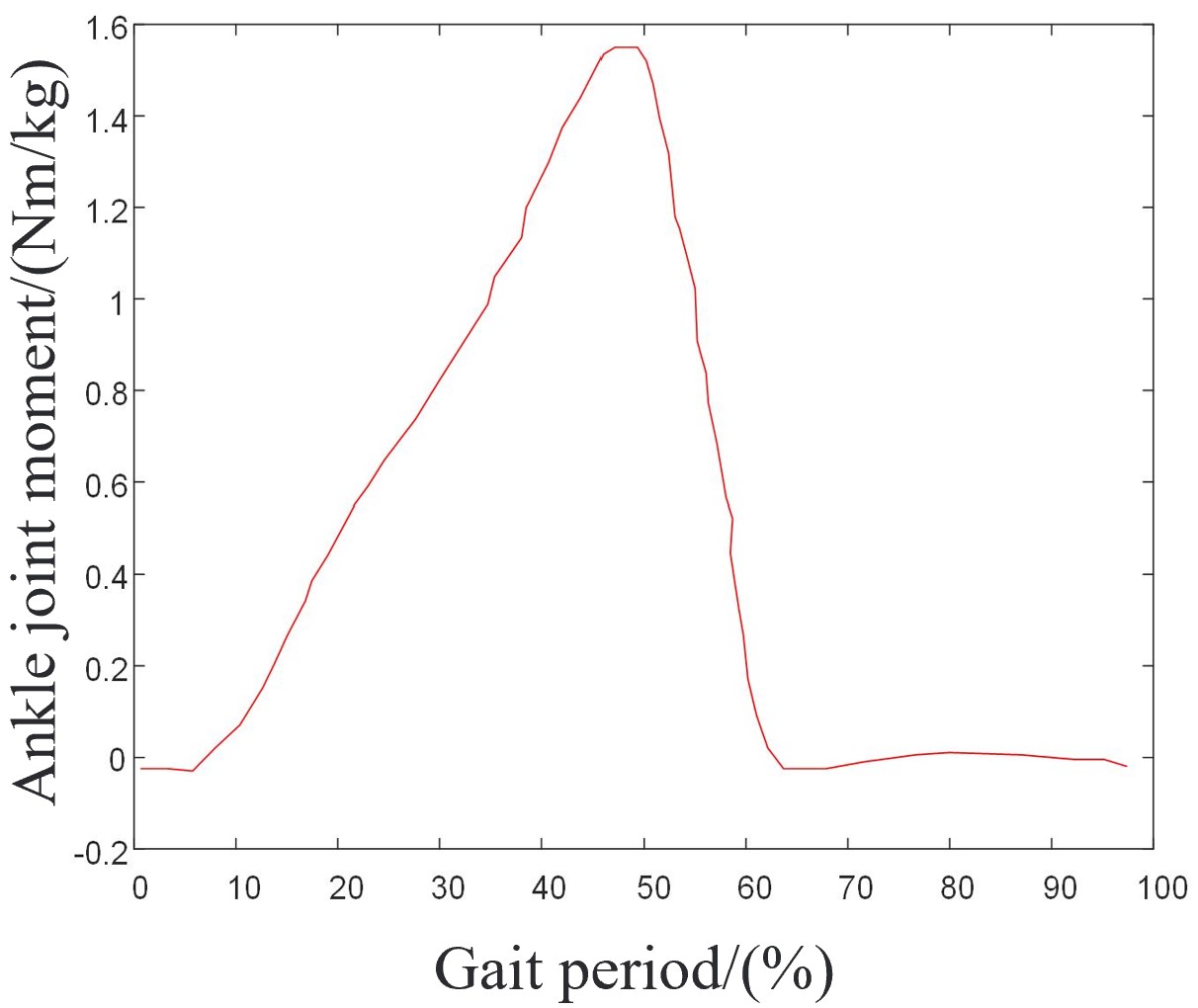}
        \centerline{\small(b)Ankle Moment}
    \end{minipage}
    \begin{minipage}[t]{0.34\linewidth}
        \centering
        \includegraphics[width=\textwidth]{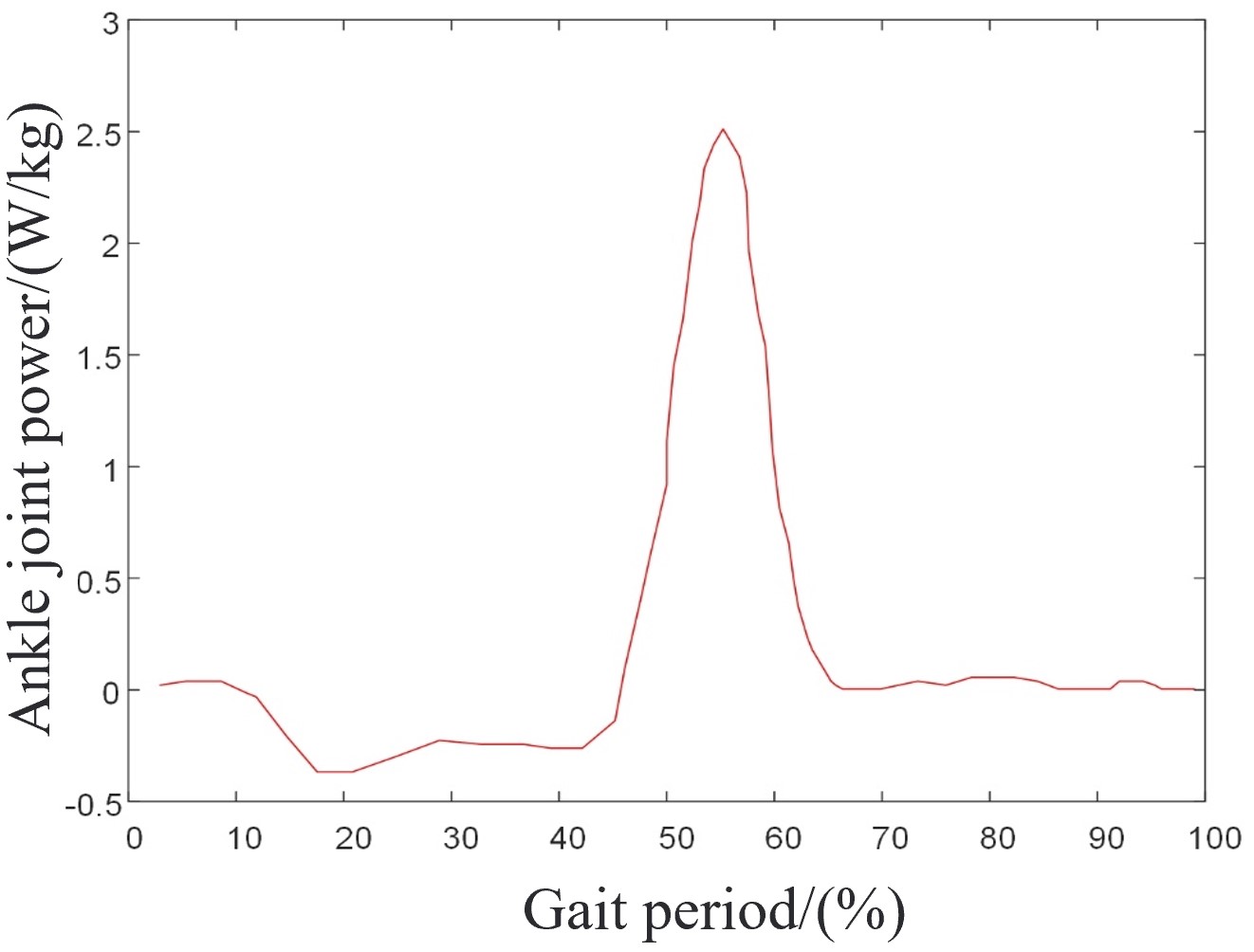}
        \centerline{\small(c)Ankle Power}
    \end{minipage}
    \caption{Ankle Joint Angle, Torque, and Power Curves during Walking}
\end{figure}

\subsection{Modeling of electro-hydraulic actuation mechanisms}
Impedance model-regulated damping is mainly carried out in two passive stages: passive plantarflexion and passive dorsiflexion.

\indent To address the aforementioned analysis, we propose the development of an electro-hydraulic driven prosthetic system that enables prosthetic actuation and energy regulation\cite{b4}. The hydraulic circuit diagram of the system is illustrated in Figure 4. 

\indent This system utilizes a low-power motor-pump unit to continuously work during the two passive phases, charging accumulator A to store energy\cite{b5}. Furthermore, it utilizes accumulator C to absorb the energy generated during the controlled dorsiflexion phase, where the ankle joint undergoes negative work due to the gravitational force acting on the human body. During the powered plantar flexion phase, the system simultaneously releases the energy stored in accumulators A and C to enhance the instantaneous power output and energy utilization efficiency.

\begin{figure}[htbp]
\centerline{\includegraphics[width=0.6\linewidth]{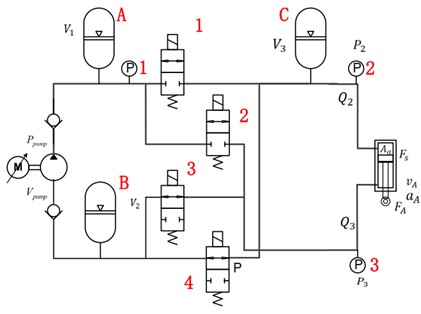}}
\caption{Hydraulic Circuit Diagram}
\label{fig}
\end{figure}

\indent Corresponding to the four phases of the human walking process, the hydraulic system of the prosthesis can also be divided into four stages: passive plantarflexion, passive dorsiflexion, active plantarflexion, and active dorsiflexion. The actuation schematic diagram is shown in Figure 5. The following is a stage-by-stage explanation of the detailed operation of each component of the system.

\begin{figure}[htbp]
\centerline{\includegraphics[width=1\linewidth]{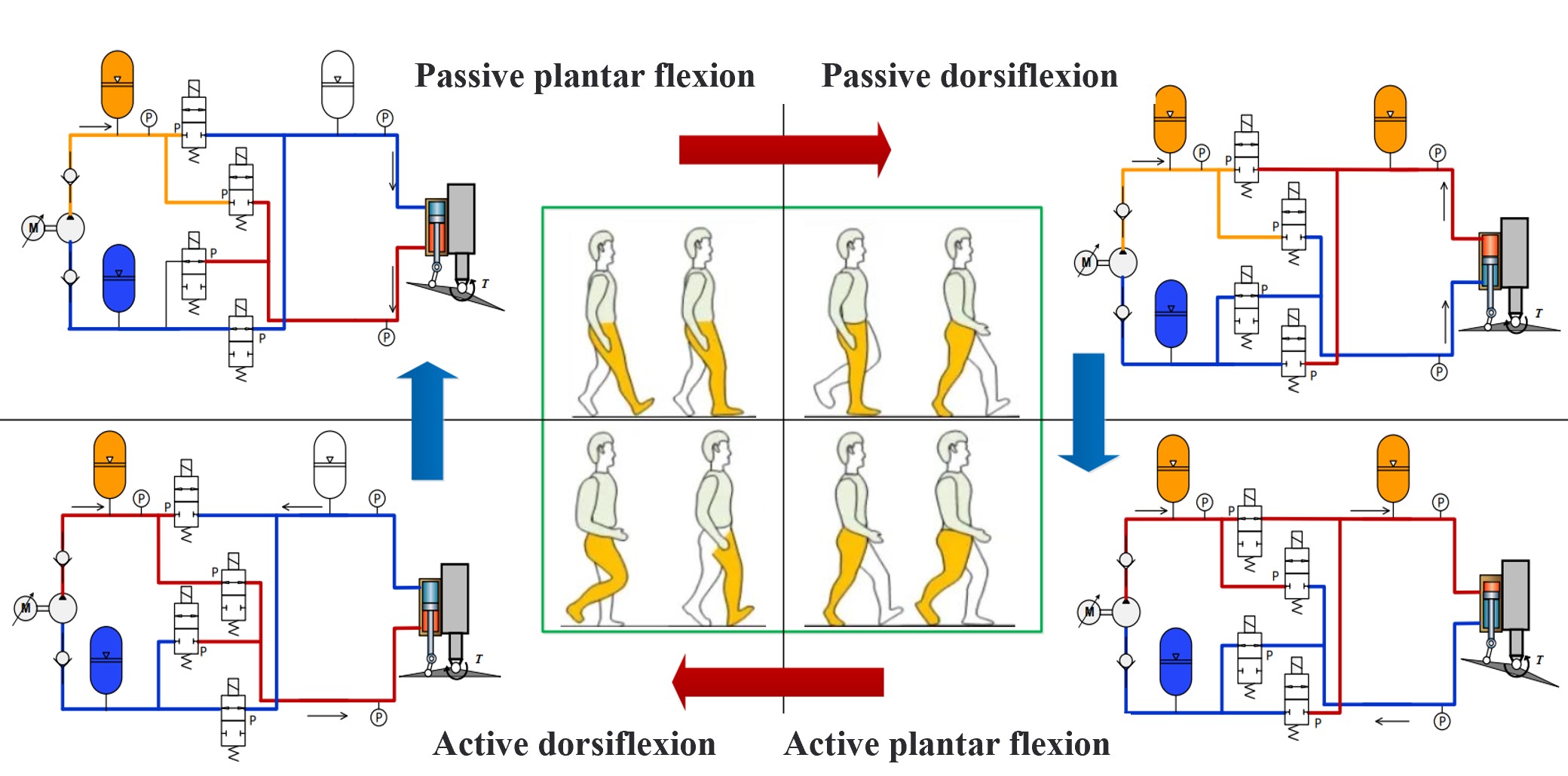}}
\caption{Prosthetic hydraulic circuit four working states}
\label{fig}
\end{figure}
\indent \textbf{The passive plantar flexion stage:} The upper and lower chambers of the hydraulic cylinder are connected through valves 3 and 4. When the heel of the foot receives an impact from the ground, hydraulic oil flows from the upper chamber to the lower chamber. At this time, the dynamic model of the hydraulic cylinder can be described as follows:
\begin{equation}
m_Aa_A=F_A-\left( P_3-P_2 \right) A_a-F_s\label{eq}
\end{equation}
\indent Where is the mass piston and load,  $x_p$ is load displacement, $P_a$ and $P_b$ is the pressure in the upper and lower chambers of the hydraulic cylinder, $F$is the external loading force,  $F_s$ is the internal friction, and $A_p$ is the area of the hydraulic cylinder in which the fluid acts.
\indent The pressure dynamic equation of the upper cylinder of the hydraulic cylinder is:

\begin{equation}
\dot{P}_a=\frac{\beta _e}{V_a-A_px_p}(A_p\dot{x}_p-Q_a)\label{eq}
\end{equation}
\indent Where $m_A$ represents the load and piston mass, $a_A$ is the acceleration of the hydraulic cylinder piston rod, $P_3$ and $P_2$ represent the pressure in the upper and lower chambers of the hydraulic cylinder, $F_A$ is the externally applied load, $F_s$ is the internal friction force, and $A_a$ is the effective area of the hydraulic cylinder.The dynamic equation for the pressure in the lower chamber of the hydraulic cylinder is given by:
\begin{equation}
\dot{P}_3=\frac{\beta_e}{V_3-A_a y_a}\left(A_a v_A-Q_3\right)\label{eq}
\end{equation}
\indent Similarly, the dynamic equation for the pressure in the upper chamber of the hydraulic cylinder is given by:
\begin{equation}
\dot{P}_2=\frac{\beta_e}{V_2+A_a y_a}\left(Q_2-A_a v_A\right)\label{eq}
\end{equation}
\indent Where $v_A$ is the velocity of the hydraulic cylinder piston, $\beta_e$ is the effective bulk modulus, $Q_3$ and $Q_2$ are the flow rates in the upper and lower chambers of the hydraulic cylinder, respectively.Meanwhile, the DC motor continues to operate, charging the high-pressure accumulator A. The volume change of the accumulator is given by:
\begin{equation}
\Delta V=V_1-V_1^{\prime}=V_0\left[\left(\frac{P_a}{P_1}\right)^{1 / n}-\left(\frac{P_a}{P_1^{\prime}}\right)^{1 / n}\right]\label{eq}
\end{equation}
\indent Where $V_1$ and $V_1'$ represent the volume of the accumulator at different times, and $P_1$ and $P_1'$ represent the pressure of the accumulator at different times. $V_0$ is the capacity of the accumulator, and $P_a$ is the charging pressure of accumulator A.

\textbf{The passive dorsiflexion stage:} The upper and lower chambers of the hydraulic cylinder are connected through valves 3 and 4. Hydraulic oil flows from the lower chamber to the upper chamber. Similar to the previous stage, the dynamic equation and pressure equation of the hydraulic cylinder are as follows:
\begin{equation}
\begin{aligned}
& m_A a_A=F_A-\left(P_2-P_3\right) A_a-F_s \\
& \left\{\begin{array}{l}
\dot{P}_2=\frac{\beta_e}{V_2-A_a y_a}\left(A_a v_A-Q_2\right) \\
\dot{P}_3=\frac{\beta_e}{V_3+A_a y_a}\left(Q_3-A_a v_A\right)
\end{array}\right.
\end{aligned}\label{eq}
\end{equation}
\indent Furthermore, the motor continues to work to charge the high-pressure accumulator A, causing the pressure in the high-pressure accumulator to increase continuously:
\begin{equation}
\begin{gathered}
\Delta V=V_1-V_1^{\prime}=V_0\left[\left(\frac{P_a}{P_1}\right)^{1 / n}-\left(\frac{P_a}{P_1^{\prime}}\right)^{1 / n}\right] \\
P_{a c c}=P_a q_{a c c}=P_a \frac{d V}{d t}
\end{gathered}\label{eq}
\end{equation}
\indent At the same time, the force of gravity acting on the hydraulic cylinder contributes to the work done, and the accumulator C recovers the energy associated with the gravitational potential energy of the human body:
\begin{equation}
\begin{gathered}
\Delta V=V_3-V_3{ }^{\prime}=V_0\left[\left(\frac{P_c}{P_3}\right)^{1 / n}-\left(\frac{P_c}{P_3{ }^{\prime}}\right)^{1 / n}\right] \\
P_{a c c}=P_c q_{a c c}=P_c \frac{d V}{d t}
\end{gathered}\label{eq}
\end{equation}

\indent Here, $V_1$, $V_1^\prime$, $V_3$, and $V_3^\prime$ represent the volumes of accumulators A and C at different times, while $P_1$, $P_1^\prime$, $P_3$, and $P_3^\prime$ represent the pressures of accumulators A and C at different times. $V_0$ denotes the capacity of the accumulators, and $P_a$ and $P_c$ represent the charging pressures of accumulators A and C, respectively.

\textbf{The active plantarflexion stage:} In this phase, the lower chamber of the hydraulic cylinder is connected to the low-pressure accumulator $\mathrm{B}$ through valve 3. At this time, the high-pressure accumulators $\mathrm{A}$ and $\mathrm{C}$ release the high-pressure oil accumulated during the passive phase. The released energy drives the hydraulic cylinder to generate an instantaneous high thrust, enabling the ankle to propel the human body forward. The dynamic equation and pressure equation of the prosthetic limb are as follows:
\begin{equation}
\begin{aligned}
& m_A a_A=F_A+\left(P_2-P_3\right) A_a-F_s \\
& \left\{\begin{array}{l}
\dot{P}_3=\frac{\beta_e}{V_3-A_a y_a}\left(A_a v_A-Q_3\right) \\
\dot{P}_2=\frac{\beta_e}{V_2+A_a y_a}\left(Q_2-A_a v_A\right)
\end{array}\right.
\end{aligned}\label{eq}
\end{equation}

\indent In this state, both the hydraulic pump and the accumulator act on the actuator. The power of the hydraulic pump is given by:
\begin{equation}
P_{\text {pump}}=P_1 V_{\text {pump}} n \eta_0 / 60\label{eq}
\end{equation}

\indent In the equation for the power of the hydraulic pump, the term $V_{\text{pump}}$ represents the displacement of the hydraulic pump.
\indent At this point, the total input power of the hydraulic system is the sum of the input power of the hydraulic pump and the power of the accumulator. The power of the accumulator comes from two sources:\\
\indent (1)The motor does work during the passive plantar flexion and passive dorsiflexion phases, storing energy in accumulator A.\\
\indent (2)The gravitational force exerted by the body does work during the passive dorsiflexion phase, recovering energy and storing it in accumulator C.\\
\indent This allows for precise control of energy storage and release, significantly improving the energy efficiency of the system.

\textbf{The active dorsiflexion stage:} The hydraulic system in this case is similar to a conventional hydraulic drive system. The motor drives the hydraulic pump, which in turn powers the hydraulic cylinder to prepare for the next gait cycle. The dynamic equation and pressure equation of the hydraulic cylinder are as follows:
\begin{equation}
\begin{aligned}
& m_A a_A=F_A+\left(P_3-P_2\right) A_a-F_s \\
& \left\{\begin{array}{l}
\dot{P}_2=\frac{\beta_e}{V_2-A_a y_a}\left(A_a v_A-Q_2\right) \\
\dot{P}_3=\frac{\beta_e}{V_3+A_a y_a}\left(Q_3-A_a v_A\right)
\end{array}\right.
\end{aligned}\label{eq}
\end{equation}

\section{Integrated and lightweight design of the overall prosthetic structure}
\subsection{Three-dimensional structural design of the ankle prosthesis}\label{AA}
The integrated design of the ankle prosthesis is shown in Figure 6. The structure consists of hydraulic integrated valve body, accumulator, piston pump, single-rod hydraulic cylinder, normally open and normally closed solenoid valves, and other hydraulic components. It also includes mechanical components such as the driving motor, ankle skeleton, and carbon fiber footplate. Additionally, it incorporates sensing elements such as liquid pressure sensors, plantar pressure sensors, and ankle angle sensors.The selection and parameters of each important component can be found in Section III(B).\\
\begin{figure}[htbp]
\centerline{\includegraphics[width=1\linewidth]{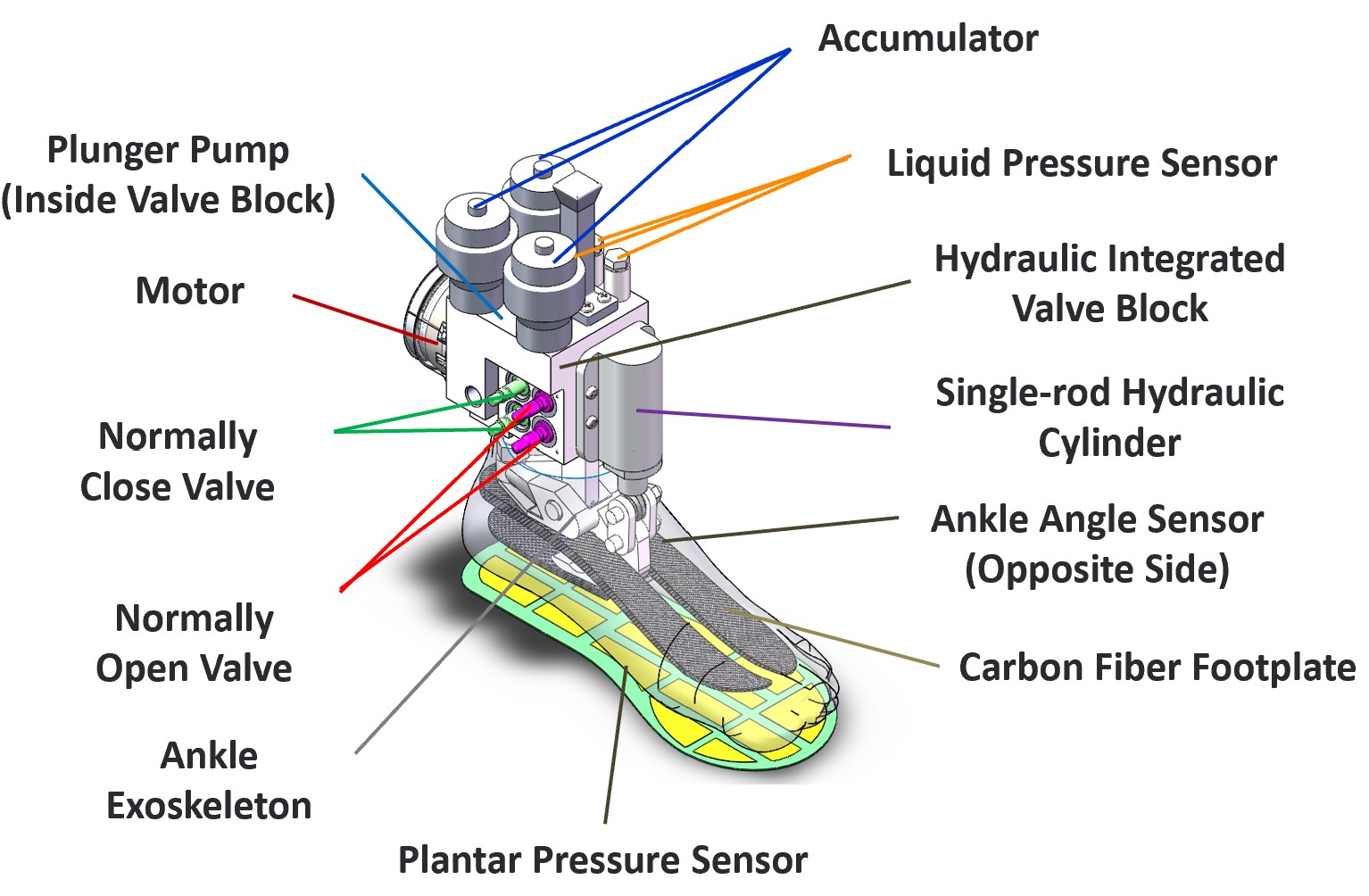}}
\caption{Structural Diagram of Ankle Prosthetic Limb}
\label{fig}
\end{figure}

\indent This ankle prosthesis incorporates an integrated internal piping design. It combines with the supporting structural body to form an integrated valve block, which allows the entire hydraulic system to be integrated onto the human ankle joint. The prosthesis achieves integration and lightweight design through techniques such as additive manufacturing-based integrated hydraulic valve block structure, optimization of internal pipeline simplification, miniaturization and integration of hydraulic components, and improvement of the ankle skeleton's topological structure. These approaches reduce the volume, mass, and moment of inertia of the prosthesis while increasing its overall power-to-weight ratio.

\subsection{Ankle prosthesis element design and selection}
A small hydraulic axial piston pump with a displacement of 65.6mm$^3$, designed pressure of 180 bar, and an approximate rated efficiency of 60\% is used in the hydraulic pump selection. It relies on the reciprocating motion of the piston in the cylinder to change the volume of the working chamber and achieve oil suction and oil pressure. The piston pump has advantages such as high rated pressure, compact structure, high efficiency, and convenient flow control, which closely match the high-pressure and high-efficiency operational requirements of the prosthetic device.

\indent For motor selection, a frameless motor TBM-6013-B from Kollmorgen is chosen with a rated power of 115W, designed voltage of 24V, and a rated speed of 3850 RPM. The frameless motor consists only of a stator and rotor, resulting in a compact size. Following the principle of simplification, the motor housing and output shaft are designed to reduce unnecessary components and lighten the overall weight of the existing mechanism. A driver board that integrates drive and magnetic encoding is used to further reduce the weight and volume of the motor components. The overall motor configuration is shown in Figure 7.
\begin{figure}[htbp]
\centerline{\includegraphics[width=0.40\linewidth]{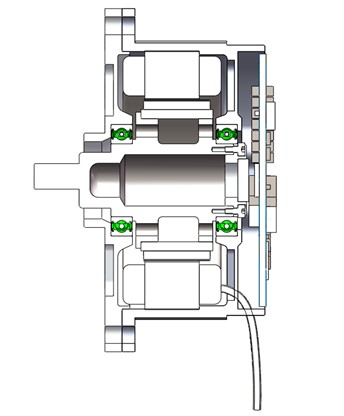}}
\caption{Structural Diagram of Prosthetic Motor}
\label{fig}
\end{figure}

\indent The hydraulic cylinder has an effective working area of 160mm$^2$, maximum actuation speed of 0.1m/s, and a maximum stroke of 40mm. The lower end of the cylinder is connected to the ankle skeleton through a rod-end joint bearing, forming a 4-bar linkage mechanism. The ankle skeleton and footplate are actuated by hydraulic pressure.
\indent The piston pump and high-speed switching valve are embedded in the prosthetic device. This helps to reduce the size of the prosthetic device, minimize overall volume and hydraulic line length, and effectively mitigate issues such as pipe vibrations and interface leaks. Refer to Figure 8 for illustration.
\begin{figure}[htbp]
\centerline{\includegraphics[width=0.55\linewidth]{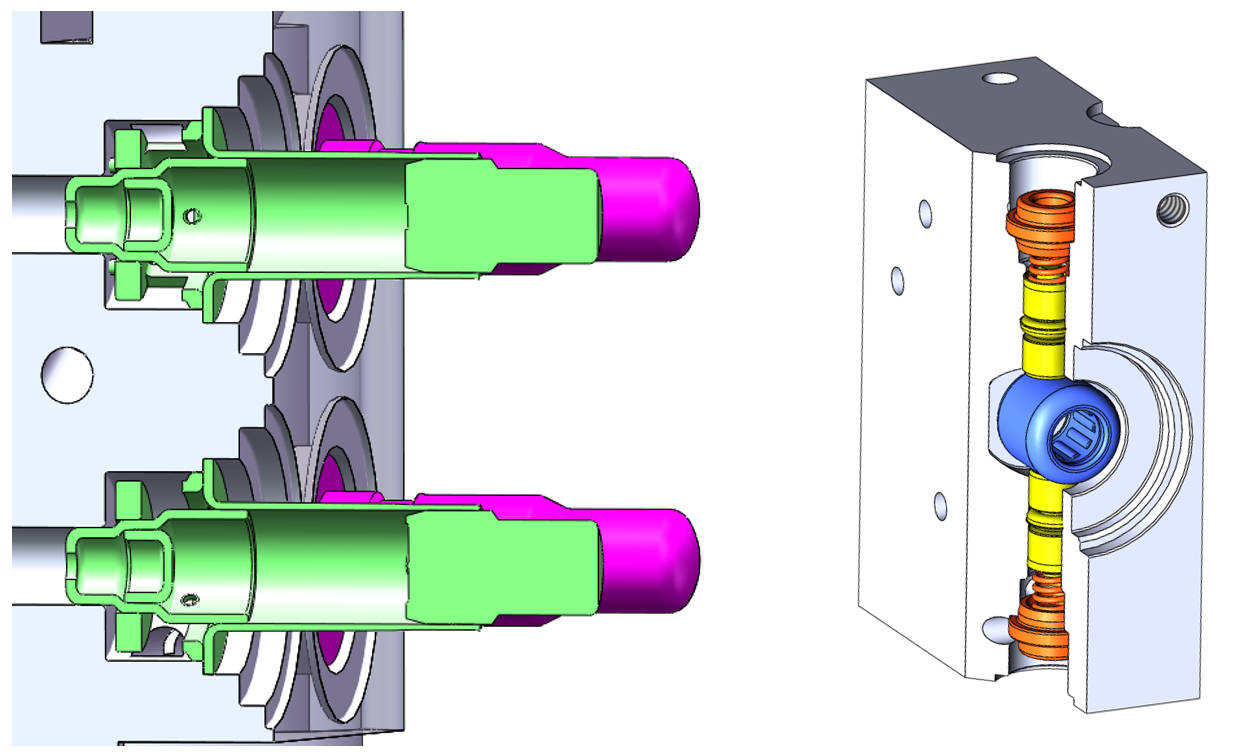}}
\caption{Integrated Design of Plunger Pump and High-Speed Switching Valve}
\label{fig}
\end{figure}
\subsection{Simulation of prosthetic output capacity}
According to the designed micro hydraulic circuit in Section II(B), an AMESim model is created. The parameters of the motor, axial piston pump, normally open valve, and accumulator are listed in the table 1. The modeling model is established based on the selected components from the hydraulic library according to the system schematic, as shown in Figure 9.
\begin{table}[htbp]
\caption{Hydraulic system simulation parameters}
\begin{center}
\begin{tabular}{|c|c|c|}
\hline
\textbf{\textit{Motor Speed}} & \textbf{\textit{Pump Displacement}}& \textbf{\textit{Pipe Diameter}} \\
\hline
3850 rpm&65.6 $mm^3$&3.2 mm  \\
\hline
\textbf{\textit{Cylinder Effective Area}} & \textbf{\textit{Valve Frequency}}& \textbf{\textit{Flow Rate}}   \\
\hline
160 $mm^2$&200 Hz&1.2 L/min  \\
\hline
\textbf{\textit{Accumulator Pressure}} & \textbf{\textit{Accumulator Capacity}}& \textbf{\textit{Value Diameter}} \\
\hline
80,70,20 bar&0.013 L&0.76 mm  \\
\hline
\end{tabular}
\label{tab1}
\end{center}
\end{table}
\begin{figure}[htbp]
\centerline{\includegraphics[width=0.6\linewidth]{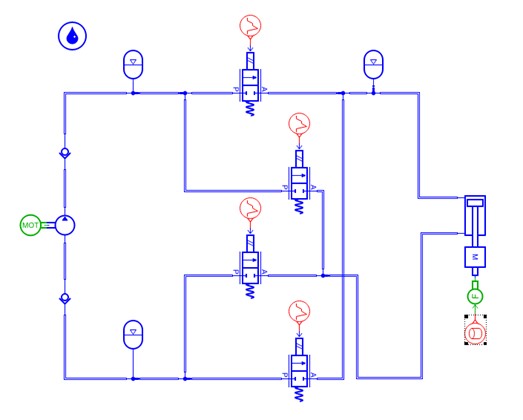}}
\caption{AMESim Model of Miniature Hydraulic Drive Circuit}
\label{fig}
\end{figure}

\indent The relationship between ankle torque and angle is influenced by walking speed, road conditions, and the physical condition of the amputee. There are two components of driving force that need to be added at the ankle joint: the driving force provided by single-rod hydraulic cylinder for the ankle joint prosthesis, and the force exerted on the ankle joint by the swinging of the lower leg during walking. The load force applied to the hydraulic cylinder is fitted based on the load force at the ankle joint during one gait cycle and is input into the system, as shown in Figure 10.
\begin{figure}[htbp]
\centerline{\includegraphics[width=0.6\linewidth]{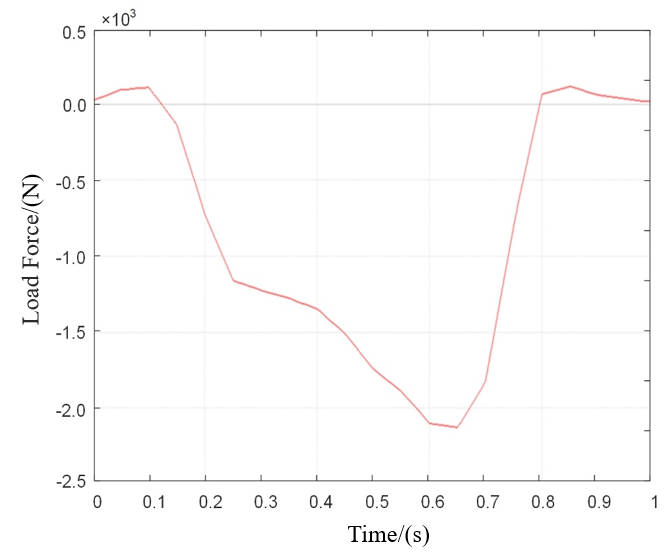}}
\caption{Load on the Ankle Joint Prosthetic Hydraulic Cylinder}
\label{fig}
\end{figure}

\indent The simulation results are shown in Figure 11, where it can be observed that the output displacement of the hydraulic cylinder in the prosthetic limb follows a similar trend to the variation of ankle angle in Figure 3. However, due to the presence of factors such as viscous friction and other power losses in the electro-hydraulic system, the final displacement of the hydraulic cylinder is slightly smaller than the design target. Nevertheless, it is sufficient to assist the human body in completing a full gait cycle.

\begin{figure}[htbp]
\centerline{\includegraphics[scale=0.23]{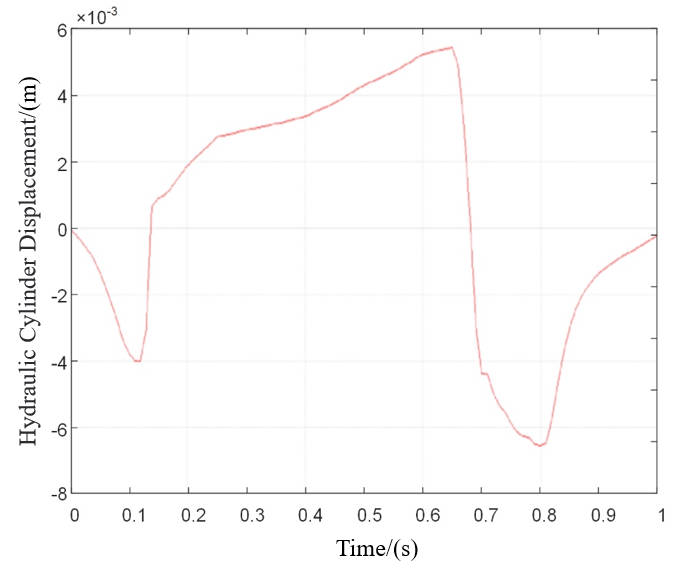}}
\caption{Output displacement of the hydraulic cylinder}
\label{fig}
\end{figure}

\section{Simulation optimization of prosthetic components}
\subsection{Optimization of Hydraulic Integrated Block Flow Channels}\label{AA}
\indent To fully utilize the advantages of additive manufacturing and achieve a minimal energy loss internal flow channel system, we conducted an analysis of the spatial flow channels. We defined two points in space that are perpendicular to the inlet and outlet axis but do not intersect, with fixed spatial coordinates. The diameter of the flow channels was set to the diameter of the prosthetic flow channels\cite{b6}. In Fluent, the outlet pressure was set to 8 MPa, and the inlet flow velocity was set to 2.5 m/s. A standard K-epsilon turbulent model was applied to complete the analysis of the complex flow channels.Through the analysis of fluid flow, we investigated the pressure loss situation of commonly used three-dimensional modeling curves to form the flow channels. The results are shown in Figure 12.\\
\indent The pressure loss results are shown in Table 2, indicating that the B-spline curve flow channel has a pressure loss of 1810 Pa under the same flow cross-sectional shape and area. It exhibits significant advantages compared to the circular radius transition flow channel and the third-order Bézier curve flow channel. Therefore, in the subsequent optimization design of the internal channels within the hydraulic valve block, B-spline curve flow channels will be generated automatically for the internal channels of the integrated block.

\begin{table}[htbp]
\caption{Three dimensional curve flow channel analysis}
\begin{center}
\begin{tabular}{|c|c|c|c|}
\hline
\textbf{Channel type} & \textbf{\textit{Inlet pressure}}& \textbf{\textit{Outlet pressure}}& \textbf{\textit{Reduction}} \\
\hline
\textbf{B-spline}&8.00169MPa&7.99988MPa&1810Pa  \\
\hline
\textbf{Circular radius}&8.00503MPa&7.99998MPa&5050Pa  \\
\hline
\textbf{Third-order Bézier}&8.00167MPa&7.99960MPa&2070Pa  \\
\hline
\end{tabular}
\label{tab1}
\end{center}
\end{table}
\begin{figure}[htbp]
    \begin{minipage}[t]{0.33\linewidth}
        \centering
        \includegraphics[width=\textwidth]{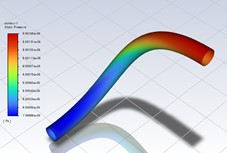}
        \centerline{\small(a)B-spline}
    \end{minipage}%
    \begin{minipage}[t]{0.32\linewidth}
        \centering
        \includegraphics[width=\textwidth]{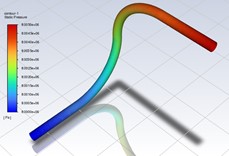}
        \centerline{\small(b)Circular radius}
    \end{minipage}
    \begin{minipage}[t]{0.33\linewidth}
        \centering
        \includegraphics[width=\textwidth]{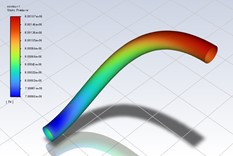}
        \centerline{\small(c)Third-order Bézier}
    \end{minipage}
    \caption{Analysis of pressure loss in the spatial flow channels}
\end{figure}

\indent The basic mathematical expression of a B-spline curve is as follows:
\begin{equation}
C(u)=\sum_{i=0}^n N_{i, p}(u) P_i\label{eq}
\end{equation}

\indent Where, $C(u)$ is the B-spline curve.$P_i$ represents the control points.$N_{i, p}$ is the $p$th order B-spline basis function.$n$ is the number of control points.The definition of B-spline basis functions can be expressed recursively. 

\indent For $p=0$ (zeroth order), it is defined as:
\begin{equation}
N_{i, 0}(u)= \begin{cases}1, & \text { if } u_i \leq u<u_{i+1} \\ 0, & \text { else }\end{cases}\label{eq}
\end{equation}

\indent For $p>0$ (orders greater than zero), it is defined as:
\begin{equation}
N_{i, p}(u)=\frac{u-u_i}{u_{i+p}-u_i} N_{i, p-1}(u)+\frac{u_{i+p+1}-u}{u_{i+p+1}-u_{i+1}} N_{i+1, p-1}(u)\label{eq}
\end{equation}

\indent Here, $u_i$ represents the $ith$ knot vector.

\indent The overall internal flow channel design process is shown in Figure 13. The optimization of the channels is based on the pressure loss at the inlet and outlet of the pipelines, aiming to reduce energy dissipation caused by local pressure loss and improve overall energy utilization efficiency. The centerline of the channel is generated based on the hydraulic schematic and the opening position, input control points, and curve order. Then, using the channel diameter and wall thickness as the profile, a scan is performed along the B-spline curve as the guiding line. The structure interference parts are modified and iterated, and the boundary conditions are determined to obtain the final valve body flow channels, as shown in Figure 13(a).

\begin{figure}[htbp]
    \begin{minipage}[t]{0.49\linewidth}
        \centering
        \includegraphics[width=\textwidth]{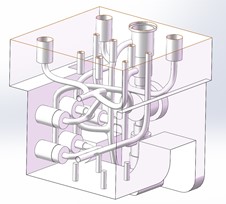}
        \centerline{\small(a)B-spline flow channel}
    \end{minipage}%
    \begin{minipage}[t]{0.49\linewidth}
        \centering
        \includegraphics[width=\textwidth]{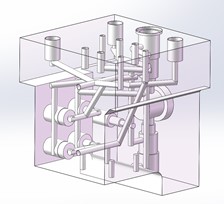}
        \centerline{\small(b)Straight flow channel}
    \end{minipage}
    \caption{Flow channel connection method based on given inlets and outlets}
\end{figure}

\indent Similarly, we analyzed the pressure loss results of the B-spline curve channels and straight channels obtained through iterative processes for the active dorsiflexion and active plantar flexion working states of the two valve bodies. The analysis results are shown in Figures 14 and 15.\\
\begin{figure}[htbp]
    \begin{minipage}[t]{0.49\linewidth}
        \centering        
        \includegraphics[width=\textwidth]{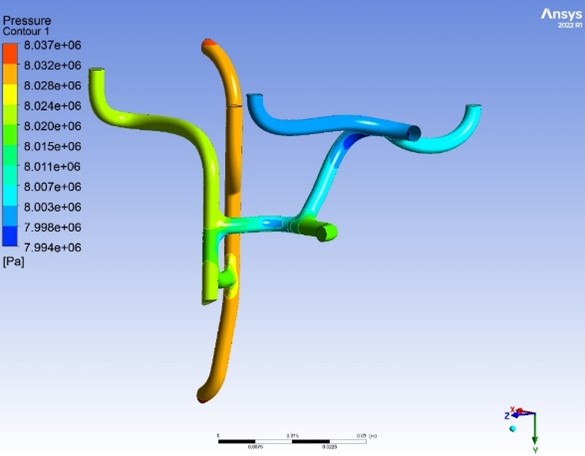}
        \centerline{\small(a)B-spline flow channel}
    \end{minipage}%
    \begin{minipage}[t]{0.5\linewidth}
        \centering
        \includegraphics[width=\textwidth]{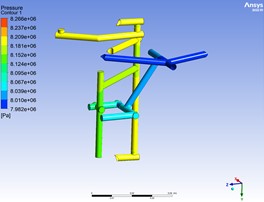}
        \centerline{\small(b)Straight flow channel}
    \end{minipage}
    \caption{Pressure distribution map during active dorsiflexion phase}
\end{figure}
\begin{figure}[htbp]
    \begin{minipage}[t]{0.49\linewidth}
        \centering
        \includegraphics[width=\textwidth]{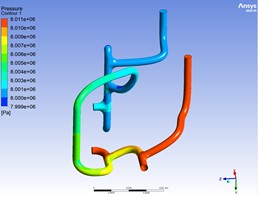}
        \centerline{\small(a)B-spline flow channel}
    \end{minipage}%
    \begin{minipage}[t]{0.50\linewidth}
        \centering
        \includegraphics[width=\textwidth]{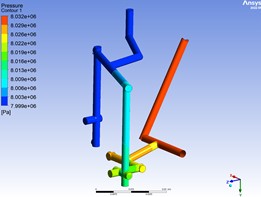}
        \centerline{\small(b)Straight flow channel}
    \end{minipage}
    \caption{Pressure distribution map during active plantarflexion phase}
\end{figure}

\indent The results are shown in Table 3. Under the same cross-sectional shape and area, the optimized channel for the plantar flexion stage has a pressure loss of 42,530 Pa, while the traditional channel has a pressure loss of 284,020 Pa, resulting in an optimization effect of 85\%. For the dorsiflexion stage, the optimized channel has a pressure loss of 12,070 Pa, while the traditional channel has a pressure loss of 33,220 Pa, resulting in an optimization effect of 64\%. Under the conditions of precision casting and additive manufacturing, the B-spline curve channels can significantly reduce the overall pressure loss.\\
\begin{table}[htbp]
\caption{Comparison of pressure loss between B-spline channel and straight channel}
\begin{center}
\begin{tabular}{|c|c|c|c|}
\hline
\textbf{Channel type} & \textbf{\textit{Inlet pressure}}& \textbf{\textit{Outlet pressure}}& \textbf{\textit{Reduction}} \\
\hline
\textbf{\makecell[c]{B-spline channel in \\ dorsiflexion phase} }&8.037MPa&7.995MPa&42530Pa  \\
\hline
\textbf{\makecell[c]{Straight channel in \\ dorsiflexion phase} }&8.266MPa&7.982MPa&284020Pa  \\
\hline
\textbf{\makecell[c]{B-spline channel in \\ plantarflexion phase} }&8.011MPa&7.999MPa&12070Pa  \\
\hline
\textbf{\makecell[c]{Straight channel in \\ plantarflexion phase}}&8.032MPa&7.999MPa&33220Pa  \\
\hline
\end{tabular}
\label{tab1}
\end{center}
\end{table}
\subsection{Lightweight optimization of ankle prosthetic components}
The weight of existing components is analyzed to identify the potential for lightweight optimization. Apart from the power components, we found that the hydraulic integrated valve block and the ankle skeleton are major contributors to the overall weight of the prosthetic system, offering significant potential for lightweighting\cite{b7}.

\indent For the hydraulic integrated block, a design approach using metal additive manufacturing is employed to achieve lightweight optimization. After component layout and flow channel optimization, the integrated block is subjected to lightweight design by removing redundant structures and internal filling to reduce its overall weight. Studies have shown that the smoothness, connectivity, and quasi-self-similarity of triply periodic minimal surfaces (TPMS) can ensure the continuous and stable transmission of internal forces without additional support\cite{b8}. Therefore, TPMS is particularly suitable for internal filling in additive manufacturing. Among various complex TPMS, the P-surface has a relatively simple structure, making it easier to replicate accurately in additive manufacturing with limited processing precision. Additionally, the geometric shape of the P-surface exhibits clear periodicity in all three spatial directions, promoting more uniform stress distribution.\\
\begin{figure}[htbp]
\centerline{\includegraphics[width=0.3\linewidth]{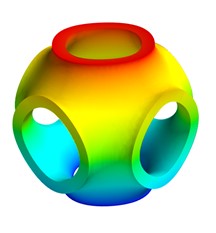}}
\caption{P-Triply
Periodic Minimal Surface}
\label{fig}
\end{figure}

\indent Therefore, in this study, three-dimensional scalar fields are used to create isosurfaces representing the polyhedron surface and filling regions. The non-essential regions inside the valve body are constructed as porous structures based on P-TPMS to reduce weight. The lightweighting problem is modeled based on the porous structure, and the multiscale porous surface with P-TPMS is constructed using an implicit function:\\
\begin{equation}
\varphi_{\mathrm{P}}=\cos (2 \pi x \cdot t(r))+\cos (2 \pi y \cdot t(r))+\cos (2 \pi z \cdot t(r))=0\label{eq}
\end{equation}
\indent Among them,$\mathrm{r}=(\mathrm{x}, \mathrm{y}, \mathrm{z})$ is a coordinate vector in the Cartesian coordinate system, and $t(r)>0$ be a periodic control function. It should be a scalar field function with a continuous distribution in space. Here, the function distribution of the periodic parameter $t(r)$ is obtained using the method of interpolation with compactly supported radial basis functions.
\begin{equation}
t(r)=\sum_{i=1}^{\mathrm{m}} \omega_i \psi\left(\left\|r-p_i\right\|+Q(r)\right)\label{eq}
\end{equation}
\indent To construct the multi-scale porous structure in the filling region inside the valve block, a volumetric distance field is used to represent the minimum distance from each point in space to the centerline of the pipeline. The grid data points in the hexahedral mesh system are used as 3D parameterization data, taking into account the position and thickness constraints of the pipeline centerline. The union boolean operation is performed on all field domains with thickness greater than the wall thickness.

\indent Due to the processing limitations of aluminum alloy additive manufacturing, the minimum wall thickness for the P-type triply periodic minimal surface unit is 0.4mm. The filling of the internal region of the valve block with P-type triply periodic minimal surface units can be achieved through boolean operations:
\begin{equation}
\phi_M=\phi_{T P M S} \cap \phi_S=\min \left({\phi}_{T P M S}, \phi_S\right)\label{eq}
\end{equation}

\indent The final hydraulic integrated valve body obtained by filling with P-TPMS is shown in Figure 17.
\begin{figure}[htbp]
\centerline{\includegraphics[scale=0.20]{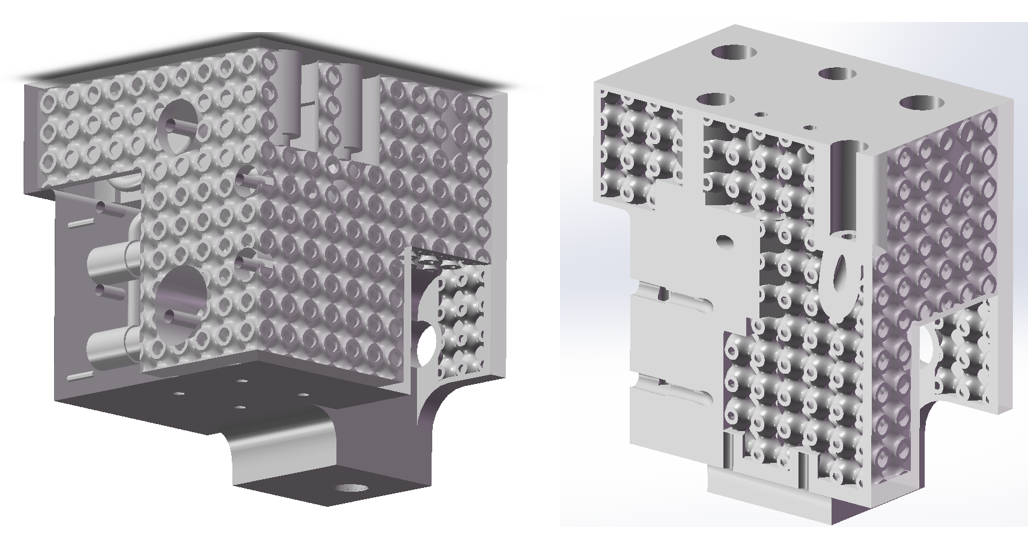}}
\caption{P-TPMS filled hydraulic integrated valve body}
\label{fig}
\end{figure}

\indent Structural strength analysis was performed on the hydraulic integrated valve body, taking into account real conditions such as human body load and ground reaction forces. The analysis results are shown in Figure 18. The maximum stress of the P-TPMS filled model is 37.971 MPa, which is lower than the yield strength of the aluminum alloy AlSi10Mg (245 MPa). The total deformation is 0.00062 mm, satisfying the structural strength requirements. Overall, the P-TPMS filled model has a mass of 51.2\% compared to the solid model, while still meeting the strength requirements. It can be concluded that the P-TPMS filling has a significant optimization effect on the lightweight design of the hydraulic integrated block.
\begin{figure}[htbp]
    \begin{minipage}[t]{0.5\linewidth}
        \centering
        \includegraphics[width=\textwidth]{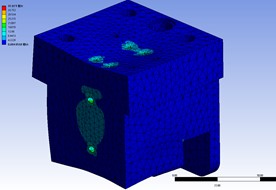}
        \centerline{\small(a)Equivalent stress diagram}
    \end{minipage}%
    \begin{minipage}[t]{0.43\linewidth}
        \centering
        \includegraphics[width=\textwidth]{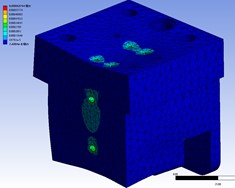}
        \centerline{\small(b)Equivalent strain diagram}
    \end{minipage}
    \caption{Stress-strain diagram of the filled model}
\end{figure}

\indent For a given ankle exoskeleton with known load conditions, constraints, and performance indicators, based on the principles of structural topology optimization, a lightweight optimization modeling framework is applied to optimize the ankle exoskeleton model\cite{b9}. The objective is to reduce its mass and minimize material consumption while ensuring the mechanical performance of the object model.

\indent The process involves defining element types, material properties, etc. Initially, as shown in Figure 19(a), loads and boundary conditions are applied to the original model to obtain equivalent stress, as shown in Figure 19(b). Subsequently, optimization initial parameters are set, including mesh size, number of elements, volume retention percentage, penalty factor, and exclusion zones. Topology optimization is then performed based on structural topology optimization theory, resulting in the optimized design shown in Figure 19(c).

\begin{figure}[htbp]
    \begin{minipage}[t]{0.33\linewidth}
        \centering
        \includegraphics[width=\textwidth]{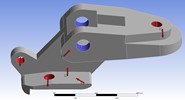}
        \centerline{\small(a)Boundary and Load}
    \end{minipage}%
    \begin{minipage}[t]{0.32\linewidth}
        \centering
        \includegraphics[width=\textwidth]{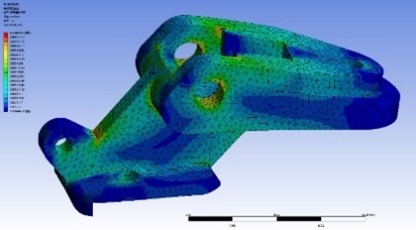}
        \centerline{\small(b)Equivalent stress}
    \end{minipage}
    \begin{minipage}[t]{0.33\linewidth}
        \centering
        \includegraphics[width=\textwidth]{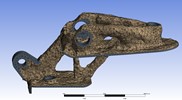}
        \centerline{\small(c)Topological Analysis}
    \end{minipage}
    \caption{Topology optimization process}
\end{figure}
\indent After iterations and optimizations, the final obtained topology result is further refined to obtain the topology reconstruction model as shown in Figure 20(b).
\begin{figure}[htbp]
    \begin{minipage}[t]{0.5\linewidth}
        \centering
        \includegraphics[width=\textwidth]{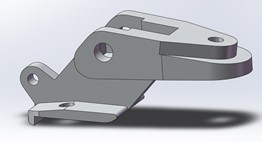}
        \centerline{\small(a)Original Model}
    \end{minipage}%
    \begin{minipage}[t]{0.5\linewidth}
        \centering
        \includegraphics[width=\textwidth]{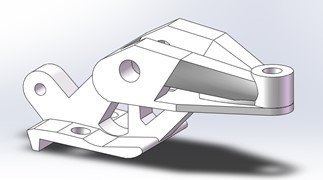}
        \centerline{\small(b)Topology Reconstruction Model}
    \end{minipage}
    \caption{Topology Optimization Result}
\end{figure}

\indent Subsequently, the structural strength of the reconstructed ankle skeleton is analyzed as shown in Figure 21. Under the same external load and boundary conditions, the maximum stress in the skeleton is 205.34 MPa, which is lower than the yield strength of aluminum alloy AlSi10Mg (245 MPa). The total deformation is 0.24 mm, meeting the requirements for strength and deformation. In this case, the mass of the reconstructed model is only 45.1\% of the original model. The lightweighting achievement is significant.
\begin{figure}[htbp]
    \begin{minipage}[t]{0.48\linewidth}
        \centering
        \includegraphics[width=\textwidth]{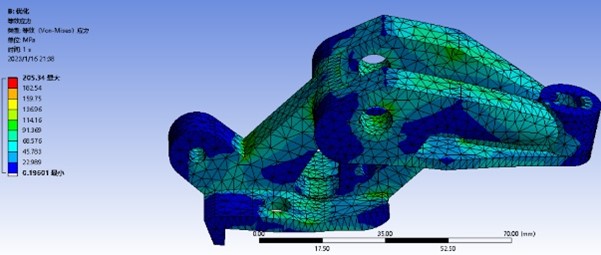}
        \centerline{\small(a)Equivalent stress diagram}
    \end{minipage}%
    \begin{minipage}[t]{0.5\linewidth}
        \centering
        \includegraphics[width=\textwidth]{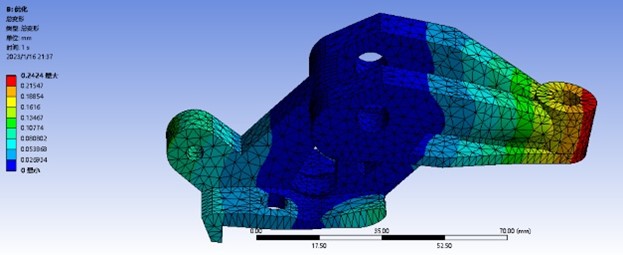}
        \centerline{\small(b)Equivalent strain diagram}
    \end{minipage}
    \caption{Equivalent Stress and Strain of Topology Reconstructed Model}
\end{figure}
\newpage 
\noindent 
\section*{Conclusion}

\indent For patients with below-knee amputation, an active ankle joint prosthesis is the most ideal compensatory tool, and scholars from various countries have conducted extensive research on it. This paper presents the design of an ankle joint prosthesis based on the principle of electro-hydrostatic actuation. The research encompasses the analysis of the hydraulic actuation mechanism and output performance, the establishment of the mechanical model and component selection, as well as the optimization of the hydraulic integrated block, pipeline, P-TMPS filling, and the topological design of the prosthetic ankle.While meeting the output requirements, the design tightly integrates all components within the smallest space possible, achieving a lightweight and integrated design of the prosthesis components, which is further validated through simulation.

\section*{Acknowledgment}

This work is supported by the Zhejiang Provincial Natural Science Foundation of China (Grant No. Z23E050032) and the National Natural Science Foundation of China (Grants No. 52275044).

\end{document}